\pdfoutput=1

\documentclass[11pt]{article}

\usepackage[]{misc/acl}

\usepackage{amsmath} 
\usepackage{graphicx}
\usepackage{hyperref}
\usepackage[capitalize]{cleveref}
\usepackage[normalem]{ulem} 

\usepackage{times}
\usepackage{latexsym}
\usepackage{booktabs}
\usepackage{enumitem}
\usepackage{float}
\usepackage{stfloats}
\usepackage{bm}
\usepackage{multirow}
\usepackage[T1]{fontenc}
\usepackage[utf8]{inputenc}
\usepackage{microtype}
\usepackage{algorithm}
\usepackage{todonotes}
\usepackage[noend]{algpseudocode}
\usepackage{xspace}

\usepackage{setspace} 

\definecolor{ReviewerColor}{rgb}{1,0.7,0.6}
\definecolor{TodoColor}{rgb}{0.8,0.9,0.8}

\title{Sentence Ambiguity, Grammaticality and Complexity Probes}

\author{Sunit Bhattacharya$^=$ \and Vilém Zouhar$^=$ \and Ondřej Bojar \\
        Charles University, Faculty Of Mathematics and Physics \\
        Insititute of Formal and Applied Linguistics \\
        \texttt{\{bhattacharya,zouhar,bojar\}@ufal.mff.cuni.cz} \\
  }

\begin{document}

\maketitle

\begin{abstract}
It is unclear whether, how and where large pre-trained language models capture subtle linguistic traits like ambiguity, grammaticality and sentence complexity.
We present results of automatic classification of these traits and compare their viability and patterns across representation types.
We demonstrate that template-based datasets with surface-level artifacts should not be used for probing,
careful comparisons with baselines should be done and that t-SNE plots should not be used to determine the presence of a feature among dense vectors representations.
We also show how features might be highly localized in the layers for these models and get lost in the upper layers.
\end{abstract}

\let\svthefootnote\thefootnote
\let\thefootnote\svthefootnote
\let\svthefootnote\thefootnote
\newcommand\blankfootnote[1]{%
  \let\thefootnote\relax\footnotetext{#1}%
  \let\thefootnote\svthefootnote%
}

\blankfootnote{$^=$Co-first authors.}
\blankfootnote{
\begin{minipage}[c]{2.5mm}
\includegraphics[width=\linewidth]{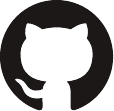}
\vspace{-1.2mm}
\end{minipage}\hspace{-1mm}
Code for the experiments in this paper is open-source:
\href{https://github.com/ufal/ambiguity-grammaticality-complexity}
{\texttt{\fontsize{7.2pt}{7.2pt}\selectfont github.com/ufal/ambiguity-grammaticality-complexity}}
}

\section{Introduction} \label{sec:intro}


Pre-trained language models, such as BERT, M-BERT \cite{devlin2019bert} and GPT-2 \cite{radford2019language}, while being very efficient at solving NLP problems, are also notoriously difficult to interpret and their analysis and interpretation is an active area of research \citep{belinkov2019analysis}.
One such technique of analysis is based on probing classifiers \citep{belinkov2021probing}, which primarily consists of training and evaluating a shallow network multi-layer perceptron (MLP) as a classifier on top of the vector representations.
Probing classifiers are now fairly established in NLP \cite{adi2016fine,tenney2019bert,ma2019universal}.

In this work, we build sentence representations from layer-wise contextual embeddings obtained from three different pre-trained language models and probe them for three linguistic traits: sentence ambiguity, grammaticality, and complexity using some well-established datasets.

In the process, we show why having a reasonable baseline is a necessity for performance interpretation.
We also demonstrate why simply visually checking the clustering of embeddings on datasets using t-SNE, a popular dimension-reduction technique in probing, can lead to incorrect conclusions.

\paragraph{Motivation.}

The study of these traits is important for example in machine translation where disambiguation is necessary and grammaticality correction and simplification sometimes happen implicitly without any control.
For the tasks of text simplification and grammar correction, it is crucial to be aware of whether and how general-purpose models encode these traits or whether they abstract the meaning from them. 
Specifically, ambiguity detection has been investigated very little in contrast to other features.
All of these three traits are orthogonal in their definitions, although their mutual relationships are unknown.
For example, it may be that ambiguous sentences tend to be more complex and prone to lower grammaticality.
We assimilate the definition of these traits from the respective datasets but nevertheless include examples in \Cref{tab:sent_examples}.

\paragraph{Contribution.}
We carry out text classification tasks of ambiguity, grammaticality and complexity and demonstrate empirically that:
\begin{itemize} 
    \item having a reasonable baseline is a necessity for performance interpretation;
    \item sentence ambiguity is represented much less than sentence complexity in the models;
    \item the template-based BLiMP dataset is not suitable for probing grammaticality because of surface-level artefacts;
    \item t-SNE is not always an adequate tool to see whether a feature is represented in vectors.
\end{itemize}

\begin{table*}[t]
\renewcommand{\arraystretch}{1.0}
\center
\resizebox{\linewidth}{!}{
\renewcommand{\arraystretch}{1.0}
\begin{tabular}{ccp{11cm}}
\toprule
\textbf{Dataset} & \textbf{Class} & \textbf{Sentence} \\
\midrule
Ambiguous COCO & Ambiguous & A metal artwork displays a clock in the middle of a floor.\\
MS COCO & Unambiguous & A couple sitting under an umbrella on a park bench.\\
\cmidrule{1-2}
HCR English & Complex & For the year, net income tumbled 61\% to \$ 86 million, or \$ 1.55 a share.\\
HCR English & Simple & In part, the trust cited the need to retain cash for possible acquisitions.\\
\cmidrule{1-2}
CoLA & Acceptable & The sailors rode the breeze clear of the rocks. \\
CoLA & Unacceptable & The problem perceives easily. \\
\cmidrule{1-2}
BLiMP-Morphology & Acceptable & The sketch of those trucks hasn't hurt Alan. \\
BLiMP-Morphology & Unacceptable & The sketch of those trucks haven't hurt Alan. \\
\cmidrule{1-2}
BLiMP-Syntax & Acceptable & Aaron breaks the glass. \\
BLiMP-Syntax & Unacceptable & Aaron appeared the glass. \\
\cmidrule{1-2}
BLiMP-Syn\_Semantics & Acceptable & Mary can declare there to be some ladders falling. \\
BLiMP-Syn\_Semantics & Unacceptable & Mary can entreat there to be some ladders falling. \\
\cmidrule{1-2}
BLiMP-Semantics & Acceptable &  There was a rug disappearing. \\
BLiMP-Semantics & Unacceptable & There was every rug disappearing.\\
\bottomrule
\end{tabular}
}
\caption{Sentence examples from used datasets.}
\label{tab:sent_examples}
\vspace{-2mm}
\end{table*}

\section{Related Work} \label{sec:related}

\paragraph{Ambiguity.}
Word-sense disambiguation has been extensively studied and is a closely related task \citep{navigli2009word}.
This has also been the focus of work done with recent NLP tools, which has mostly concentrated on the determination of ambiguity at the lexical level and not at the sentence level. 
\citet{yaghoobzadeh2019probing,csahin2020linspector,meyer2020modelling} classify ambiguous words.
\citet{chen2020probing} explore the geometry of BERT and ELMo \citep{peters-etal-2018-deep} using a structural probe to study the representational geometry of ambiguous sentences.
\citet{bordes2019incorporating} use a combination of visual and text data to ground the textual representations and make notes on disambiguation.
Ambiguity modelling has also been a focus of the MT community because translation often requires disambiguation.
This applies on many levels: lexical \citep{higinbotham1991resolution,zou2017understanding,do2020resolving,campolungo2022dibimt}, syntactic \citep{pericliev1984handling} and semantic \citep{baker1994coping,stahlberg2022jam}.
Psycholinguists have also studied the effect of ambiguity resolution on cognitive load \citep{altmann1985resolution,trueswell1996role,papadopoulou2005reading}, often motivated by issues in MT \citep{sammer2006ambiguity,scott2018translation}. 
\citet{sunit2022et} explore ambiguity by the task of translation by human annotators.

\paragraph{Grammaticality.}
This trait has been studied historically from the perspective of human sentence processing and acceptability \citep{nagata1992anchoring,braze2002grammaticality,mirault2020time}.
Many real-world applications utilize tools for automatic grammaticality prediction \citep{heilman2014predicting,warstadt2019neural}, such as automatic essay assessment \citep{foltz1999automated,landauer2003automatic,dong2017attention} or machine translation \citep{riezler2006grammatical}.
For MT, output acceptability, or fluency, is a standard evaluation direction for which many automated metrics exist \citep{hamon2006x,lavie2009meteor,stymne2010using}.
In contrast to our supervised classifier approach, perplexity-based approach has been used to measure acceptability \citep{meister2021revisiting}.

Related more closely to our setup, \citet{hewitt2019structural} use a linear probe and identify syntax in contextual embeddings.
\citet{lu2020vgcn,li2021bert} examine gramaticality in BERT layers.
\citet{hanna2021fine} assess BERTScore effectiveness in spotting grammatical errors.



\paragraph{Complexity.}

Similarly to other traits, complexity was first studied in the human processing of language \citep{richek1976effect,just1996brain,heinz2011sentence}.
\citet{brunato2018sentence} perform a crowd-sourcing campaign for English along with an in-depth analysis of the annotator agreement and complexity perception.
Automatic complexity estimation is vital, especially in the educational setting for predicting readability \citep{mcnamara2002coh,weller2020you}.
\citet{ambati2016assessing} estimate sentence complexity using a parser while \citet{vstajner2017automatic} do so using n-grams.
\citet{sarti-2020-interpreting,sarti2021looks} juxtapose the effect of complexity on language models and human assessment thereof.
\citet{martinc2021supervised} survey multiple neural approaches to complexity estimation, including using pre-trained LM representation.
In contrast to our work, they report only the final results and do not investigate the issue from the perspective of probing (e.g. what representation to extract and from which layer).

\paragraph{Probing.}
Earlier probing studies have shown that the early layers of BERT capture phrase-level information and the later layers tend to capture long-distance dependencies \citep{jawahar2019does}.
The syntax is also captured more in the early layers of BERT and higher layers are better at representing semantic information \citep{tenney2019bert}.
It is not clear if and how pre-trained models achieve compositionality \citep{kalchbrenner2013recurrent,nefdt2020puzzle,kassner2020pretrained} and how linguistic knowledge is represented in sentence embeddings.
\citet{liu2019linguistic} use probing on a set of tasks including token labelling, segmentation and pairwise relation extraction to test the abilities of contextual embeddings.
Mutual information can be used as a viable alternative to traditional probes that require optimization \citep{pimentel2020information}.
A conceptual follow-up is $\mathcal{V}$-information \citep{hewitt2021conditional} which is better suited for probing.
In many cases, t-SNE is the prevalent method of visualization of class clusters in high-dimensional vector space \citep{jawahar2019does,jin2019probing,wu2020probing,hoyt2021probing}.

\section{Data} \label{sec:data}

For each trait, we use a different dataset.
Their overall sizes are listed in \Cref{tab:data_size} and example sentences in \Cref{tab:sent_examples}.
We repurpose the datasets and derive binary labels (positive/negative) from each: ambiguous/unambiguous, complex/simple and grammatical/ungrammatical.

\vspace{2mm}

\paragraph{Ambiguity.} We use sentences from the MS COCO \cite{lin2014microsoft} dataset, for our list of ambiguous and unambiguous sentences. The MS COCO dataset comprises of a set of captions describing an image. Captions containing ambiguous verbs corresponding to 461 images (Ambiguous COCO; \citealp{elliott-EtAl:2016:VL16}) constitute the ambiguous sentences for our experiment. 461 captions that were randomly sampled from MS COCO constituted the unambiguous sentences for the experiment.

\paragraph{Complexity.}
Corpus of Sentences rated with Human Complexity Judgments\footnote{English sentences were taken from the Wall Street Journal section of the Penn Treebank.
Italian sentences were taken from the newspaper section of the Italian Universal Dependency Treebank.} \citep{iavarone2021sentence} and PACCSS-IT \citep{brunato2016paccss} contain 20 human ratings on the scale from 1 (not complex) to 7 (very complex) about sentences.
We binarize these ratings and consider sentences below the average to be simple sentences and others to be complex sentences.
The resulting dataset is class-balanced (complex/simple) in terms of examples (592 sentences of each class for English and 551 sentences for Italian).
The average sentence length for complex and simple examples is 24.84 and 13.95, respectively for English sentences.
For Italian sentences, the average sentence length for complex and simple examples is 21.61 and 12.26, respectively.
The complexity could therefore be encoded solely in the sentence length.

\paragraph{Grammaticality.}
For experiments under this category, we use the Benchmark of Linguistic Minimal Pairs (BLiMP; \citealp{warstadt2020blimp}) and the Corpus of Linguistic Acceptability (CoLA; \citealp{warstadt2019neural}) datasets.
BLiMP contains sentence pairs, one of which contains a mistake in syntax, morphology, or semantics while the other is correct.
The dataset covers 67 different conditions, grouped into 12 phenomena.
These phenomena are further categorized as `syntax',`morphology',`syntax-semantics' and `semantics'.
The CoLA dataset is not contrastive but contains human annotations of acceptable grammaticality.

\begin{table}[t]
\center
\begin{tabular}{llcc}
\toprule
& \textbf{Dataset} & \textbf{Sentences} \\
\midrule
Ambiguity & COCO & 0.9k \\
\cmidrule{1-1} 
Complexity & HCR English & 1.2k \\
           & PACCSS-IT & 1.1k \\
\cmidrule{1-1} 
Grammaticality & CoLA  & 5k \\
               & BLiMP & 67$\times$2k\\
\bottomrule
\end{tabular}
\caption{Number of sentences for each dataset corresponding to each trait.}
\label{tab:data_size}
\end{table}


\section{Experiments}

\subsection{Task definition}

\begin{figure}[htbp]
\includegraphics[width=\linewidth]{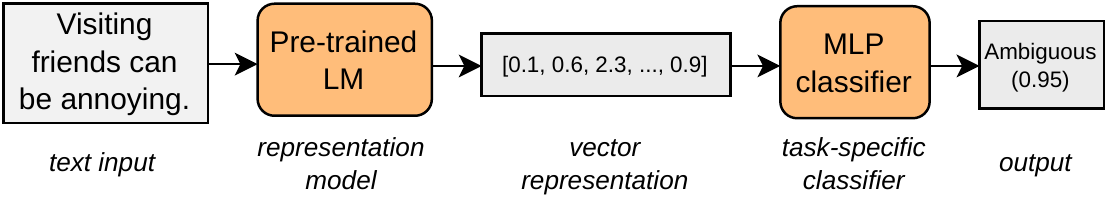}
\caption{Example of the experiment pipeline for ambiguity classification. Ambiguous sentence from \citet{stanley2000quantifier}.}
\label{fig:task_pipeline}
\end{figure}

In the following experiments, we are solving three classification tasks in parallel.
The input is always the whole sentence and the output one of the two classes (ambiguous/unambiguous, complex/simple, acceptable/unacceptable), as shown in \Cref{tab:sent_examples}, applies to the whole sentence.
The whole pipeline is also depicted in \Cref{fig:task_pipeline}.
When using the TF-IDF feature extractor, it replaces the \emph{pre-trained LM} block.

\subsection{Setup}

We use a simple MLP classifier to identify three linguistic traits from BERT (bert-base or multilingual bert-base) and GPT-2.
The resulting vectors are 768-dimensional.\footnote{The CLS and pooling representations apply only to BERT.} Both of these models are Transformer based models and contain 12 layers, which makes comparison convenient.  
We perform probing on each model separately.

\medskip

\begin{itemize} 
\item \textbf{CLS}: single vector at the \texttt{[CLS]} token.
\item \textbf{Pooling:} single vector from the pooling layer.
\item \textbf{Tokens:} vector representations of tokens aggregated with mean or (Hadamard) product to get a single 768-dimensional vector.
\end{itemize}

We obtain the layer-wise pre-trained model representations using Huggingface \cite{wolf2019huggingface} and use them to train a classifier that identifies if a sentence belongs to the positive class (e.g. ambiguous) or not.
We perform a 10-fold cross-validation each with 10 runs of MLP.

\paragraph{Baseline.}

The most common class classifier (50\% accuracy) is a poor baseline because it may be that the ambiguous and non-ambiguous sentences are distributed differently w.r.t. topic.
In an attempt to alleviate this issue, we, therefore include as the baseline a TF-IDF-based vectorizer (with a varying number of maximum features).
Probe performance of e.g. 65\% would be considered at the first glance a positive result compared to 50\%.
However, in reality, it would be a false positive finding if a simple lexical feature extractor such as TF-IDF could yield 70\%. 

\paragraph{MLP Configuration.}

For probing we use \texttt{MLPClassifier} from scikit-learn 1.1.0 \citep{scikit-learn} with most defaults preserved, as shown in \Cref{tab:mlp_params}.

\begin{table}[ht]
\centering
\begin{tabular}{cc}
\toprule
Architecture & Single hidden layer (100) \\
Activation & ReLU \\
Optimizer & Adam \\
Learning rate & $10^{-3}$ \\
Epochs & Early stopping, patience 1 \\
\bottomrule
\end{tabular}
\caption{MLP classifier configuration.}
\label{tab:mlp_params}
\end{table}

\subsection{Ambiguity \& Complexity}

\begin{figure}[t]
\includegraphics[width=\linewidth]{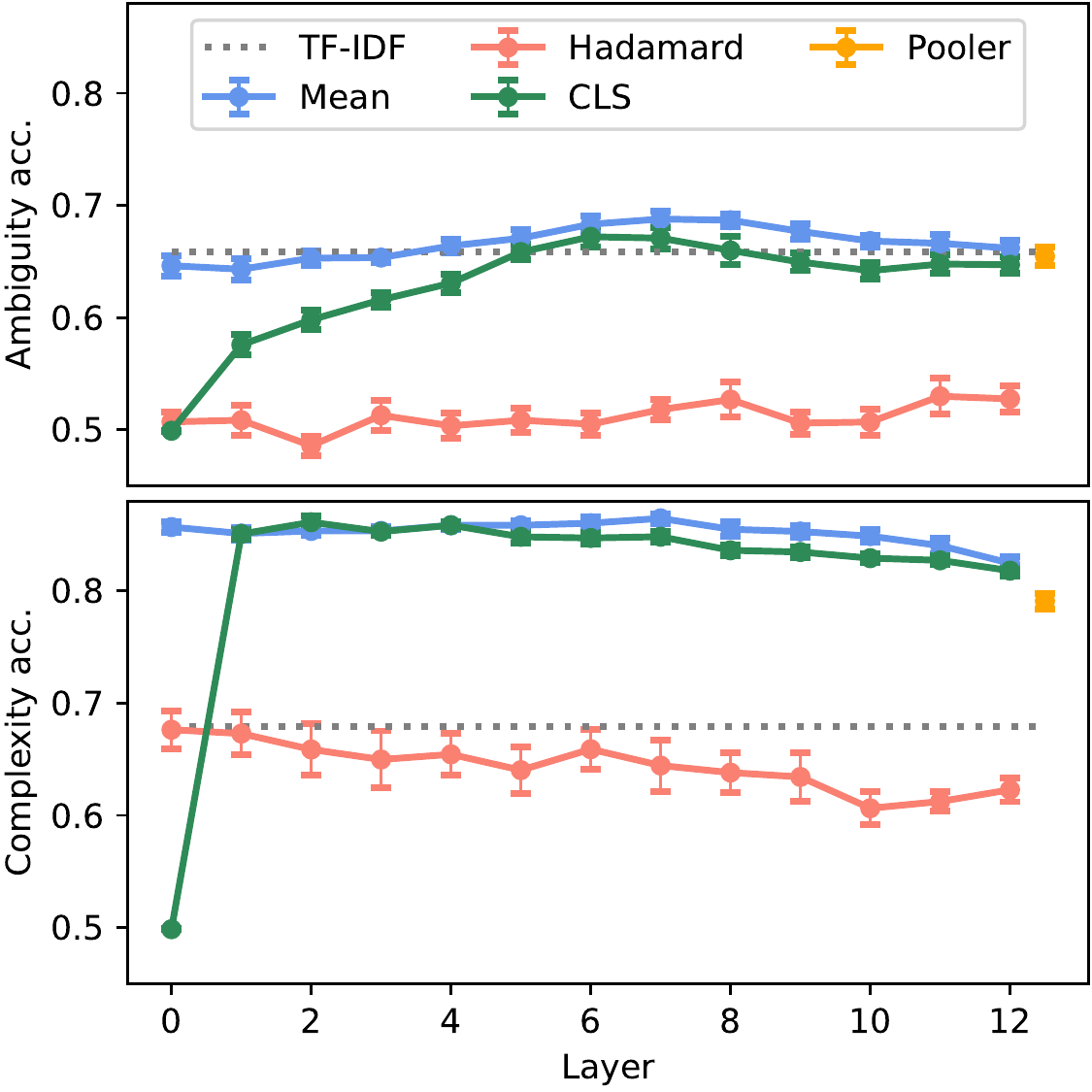}
\caption{MLP dev accuracy for \emph{ambiguity} and \emph{complexity} BERT representation across layers.}
\label{fig:amb_comp_double}
\end{figure}

Because the dataset is in Italian, we make use of multilingual BERT for both Complexity datasets.
The probe performance for M-BERT is shown in \Cref{fig:amb_comp_double}.
At the first glance, it appears that the model does represent ambiguity internally since the ambiguity probe is systematically higher than 50\%.
However, because TF-IDF performs similarly and only uses surface-level features, the probe is very weak.
This is supported by the fact that the most negative tokens from the classification (extracted from logistic regression coefficients) contained words such as \emph{man} or \emph{woman}, which disambiguate, based on gender, some unclear cases with an unclear referent.

In contrast, the complexity probe is systematically higher than the TF-IDF baseline.
With minor exceptions, the accuracy remains high regardless of the layer.
The performance for Italian (sentences taken from PACCSS-IT corpus) is identical to that for English using M-BERT (not shown).
The CLS representation at layer 0 is 50\% in both instances because it does not contain any information from the sentence (before the self-attention block).

\subsection{Grammaticality}

\begin{figure}[t]
\includegraphics[width=\linewidth]{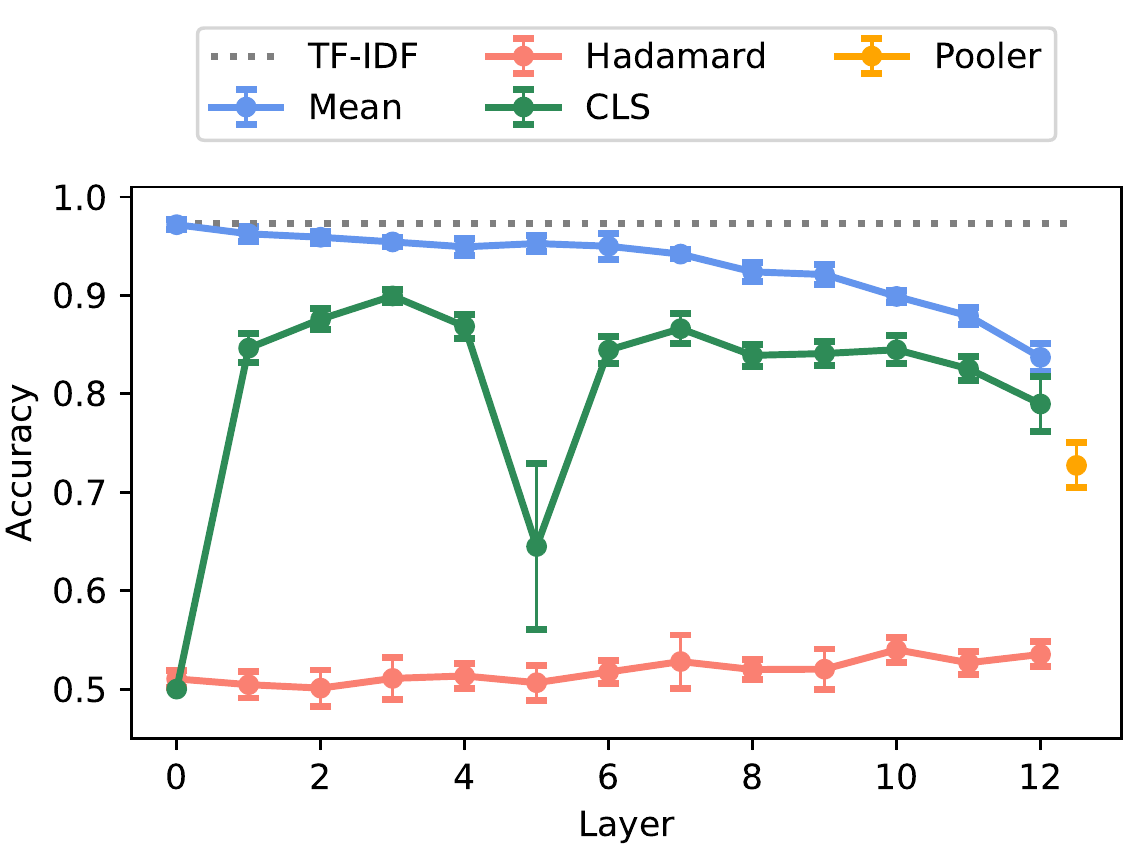}
\caption{MLP dev accuracy for \emph{determiner noun agreement irregular 1} task of BLiMP benchmark for BERT representation across layers. Each point is represented with a mean across 10 runs with a 95\% confidence interval.}
\label{fig:blimp_morphology}
\end{figure}

For the morphological task of determiner-noun agreement, \Cref{fig:blimp_morphology} shows a sudden drop in accuracy for the CLS representation at the 5th layer.
In all the tasks concerning ``Determiner-Noun Agreement'', the sentence minimal pairs focus on the number agreement between the demonstrative determiners (like this/these) and an associated noun. Examples of minimal pairs from the different tasks of this kind are shown in \Cref{tab:determiner-noun-agreement}.

While the cause is unclear, it corresponds to the average norm of the representation being very low at that particular layer, making it harder for the classifier optimization.

\begin{table*}[t]
\center
\resizebox{0.9\linewidth}{!}{
\renewcommand{\arraystretch}{1.0}
\begin{tabular}{p{7cm}p{7cm}p{2.7cm}}
\toprule
\textbf{Acceptable Sentence} & \textbf{Unacceptable Sentence} \\
\midrule
Raymond is selling this sketch. & Raymond is selling this sketches. \\
Carmen hadn't shocked these customers. & Carmen hadn't shocked these customer.\\
\midrule
Carl cures those horses. & Carl cures that horses. \\
Sally thinks about that story. & Sally thinks about those story. \\
\midrule
Laurie hasn't lifted those cacti. & Laurie hasn't lifted those cactus. \\
The waitresses haven't cleaned this thesis. & The waitresses haven't cleaned this theses. \\
\midrule
The teachers are running around this concealed oasis. & The teachers are running around these concealed oasis. \\
Randolf buys those gray fungi. & Randolf buys that gray fungi. \\
\midrule
Cynthia scans these hard books. &Cynthia scans this hard books. \\
Jerry appreciates this lost report. & Jerry appreciates these lost report. \\
\bottomrule
\end{tabular}
}
\caption{Example minimal sentence pairs from the \emph{determiner-noun} agreement task of BLiMP.}
\label{tab:determiner-noun-agreement}
\vspace{-2mm}
\end{table*}

\begin{figure}[t]
\includegraphics[width=\linewidth]{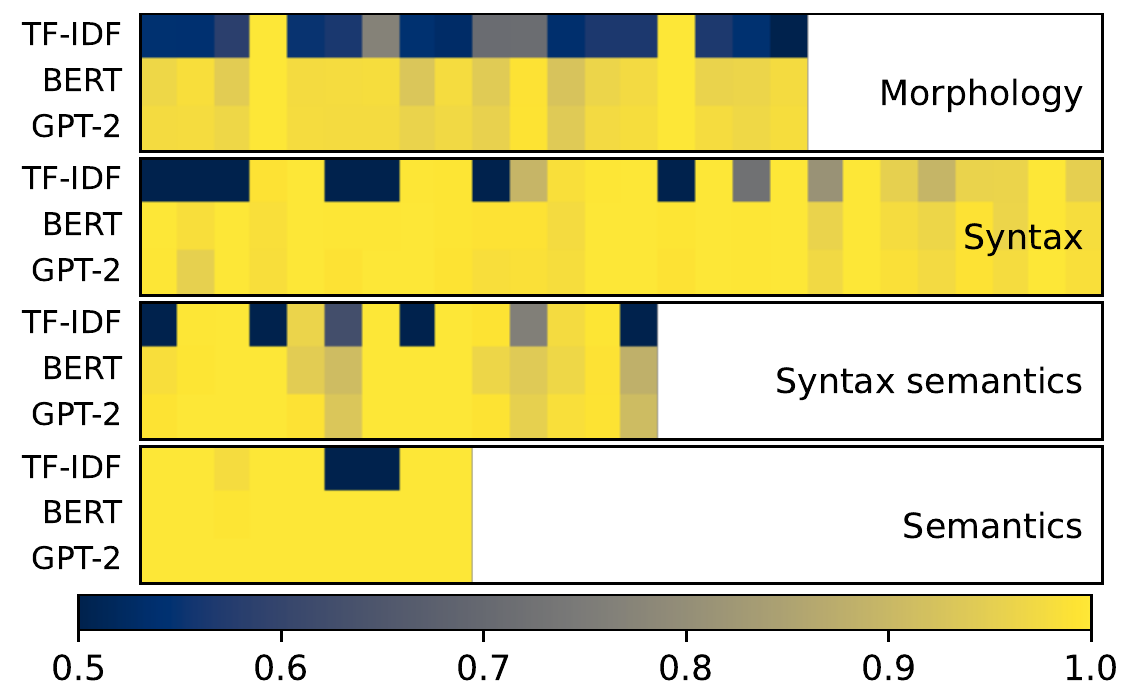}
\caption{Accuracy on various BLiMP tasks with a max of BERT and GPT-2 representations and TF-IDF baseline. Each task+model is represented as one square. The lighter squares correspond to greater accuracy and are hence better.}
\label{fig:blimp_aggregation}
\end{figure}

As \Cref{fig:blimp_aggregation} shows, many tasks can be ``solved'' with a simplistic TF-IDF featurizer, making them inadequate for determining the usefulness of large model representations.
More adequate datasets need to be developed for probing stronger models.
Systematically for all cases in morphology where the TF-IDF failed to work accurately, the performance of CLS representations was worse than the mean representations.
Even in most semantics tasks, TF-IDF probes had near-perfect accuracy.
For the 7 out of 26 syntactic tasks where the TF-IDF classifier was not accurate, the BERT models show a steep rise in accuracy from the 2nd/3rd layer for the mean and CLS representations, respectively.
In comparison, GPT-2 does not exhibit this pattern.

\section{Discussion}

The experiments with ambiguity reveal that the representations of the pre-trained models do not encode the ambiguity trait well.
The description detailing how the Ambiguous COCO was created \citep{elliott2017findings} states that the dataset was created with the intention of testing the capabilities of multimodal translation systems.
We posit that ambiguity as a trait is not encoded in an accessible way in the layer representations of pre-trained models.

For BLiMP tasks related to morphology and syntax-semantics, the accuracy goes down in the upper layers, presumably because of increasing abstraction for both models (not shown in graphs).
Although we perform experiments without fine-tuning, the findings are in line with the experimental results of \citet{mosbach2020interplay} where finetuning on 3 tasks from the GLUE benchmark \citep{wang2018glue} showed changes in probing performance mostly in the higher layers. 
Fine-tuning however led to modest gains. The present setup which probes sentence representations from pre-trained models shows that the middle layers fare far better in our probing tasks than the upper layers. This leads us to posit that the features of interest are highly localized and are lost in the upper layers (even with fine-tuning). 

Although both BERT and GPT-2 employ the Transformer \citep{vaswani2017attention} architecture, they have very different ways and locations for storing knowledge in their internal representations \cite{rogers2020primer,vulic2020probing,lin2019open,kuznetsov2020matter,de2020good,liu2021gpt}.
The CLS representations outperform the mean representations in only a few cases.
This is expected since without fine-tuning the CLS token in BERT is trained to be used for the next sentence classification tasks.

\begin{figure}[ht]
\includegraphics[width=\linewidth]{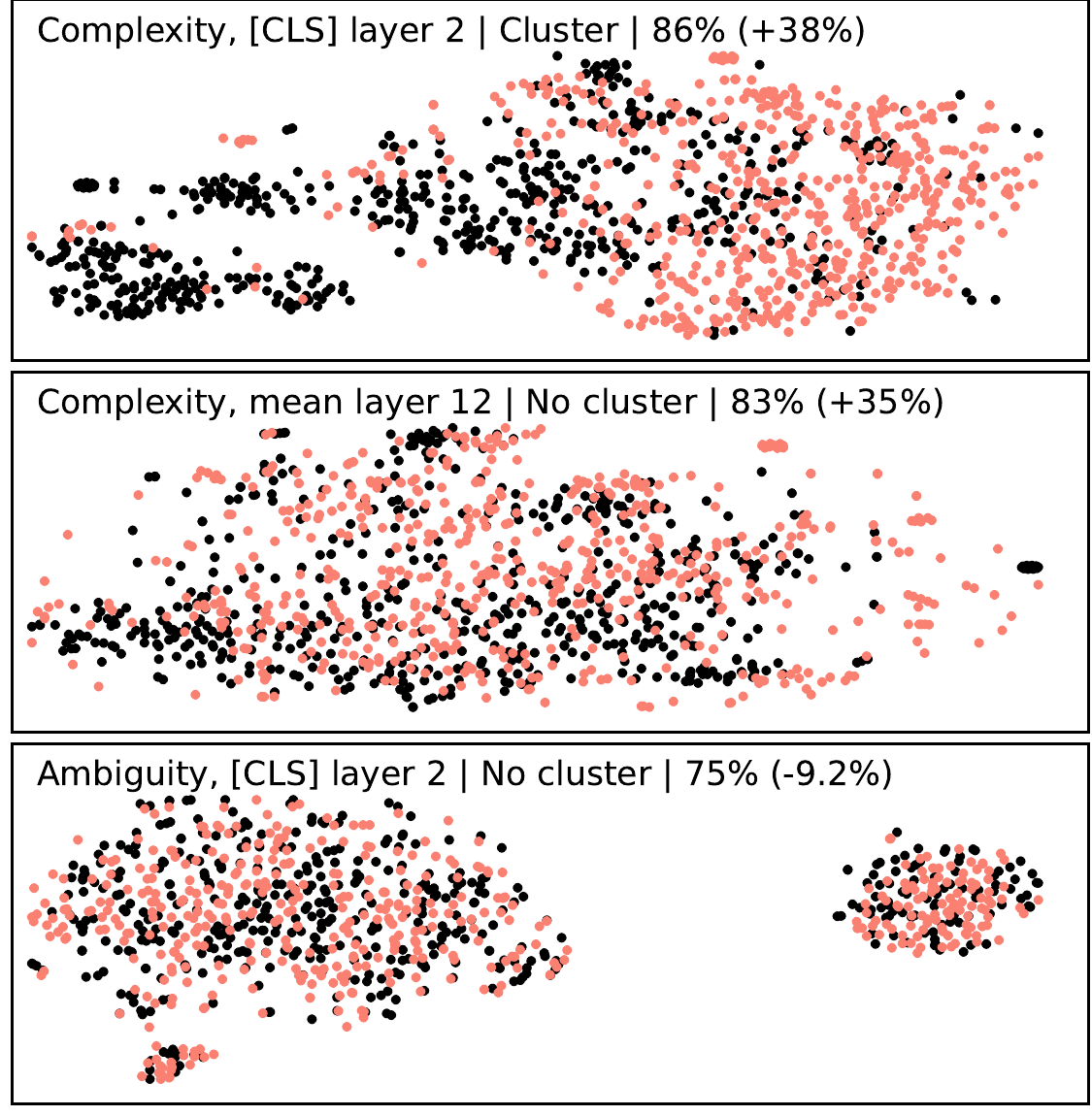}
\caption{
t-SNE projections from BERT-based embeddings. The first and the second row show high accuracy. The second and third rows show a lack of visual clusters. Red/black represent either complex/simple or ambiguous/unambiguous sentences. Percentages include classifier accuracy with the difference to the TF-IDF baseline in parentheses.}
\label{fig:tsne_triple}
\end{figure}

\begin{figure}[!htbp]
\includegraphics[width=\linewidth]{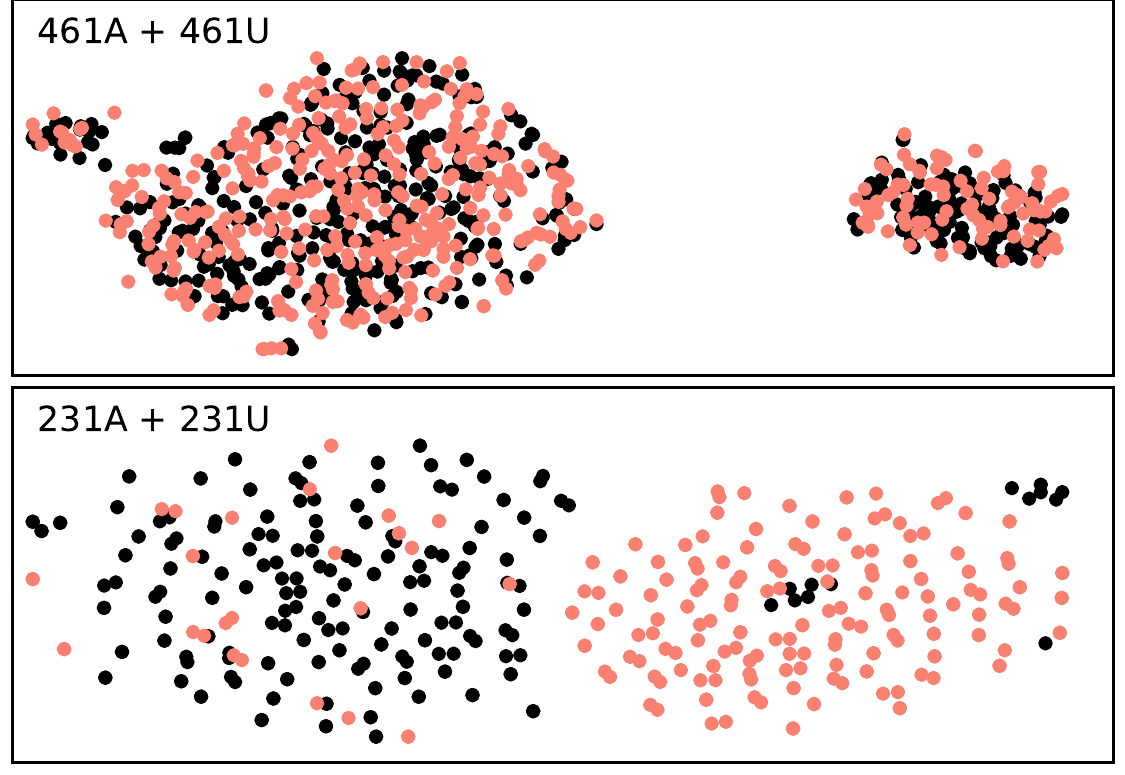}
\caption{t-SNE projections from BERT-based embeddings (layer 1 of CLS) on ambiguous/unambiguous sentences (58\% MLP and 66\% TF-IDF accuracy). The first row is all the vectors and the second is half of them subsampled by \Cref{alg:force_tsne}. Red/black represent ambiguous/unambiguous sentences.}
\label{fig:tsne_force}
\end{figure}

\section{t-SNE Inadequacy}

Given appropriate optimization and classifier, if two or more classes in a vector space form clusters, they are linearly separable and therefore the classifier performs well.
Furthermore, if a classifier probe performs well and is not affected by surface-level phenomena, it means that the features are represented in the vectors.
Both these statements are one-way implications:
\begin{itemize}
\item clear clustering $\rightarrow$ high classifier accuracy
\item high classifier acc. $\rightarrow$ feature present
\end{itemize}

Because t-SNE projects vectors from high dimensional space to lower dimensions in a manner that tries to preserve distances, it may be that visual clusters are created where there were none before and vice versa.
The following scenarios are possible:
\begin{itemize}
\item clear clusters and high classifier accuracy 
\item no clusters and high classifier accuracy 
\item no clusters and low classifier accuracy 
\end{itemize}

The last combination, ``clear clusters and low classifier accuracy'' is impossible with proper optimization.
The three scenarios on probes from the previous experiments are shown in \Cref{fig:tsne_triple}.
The conclusion is that probes should always precede visual clustering checks using t-SNE because it may be that the data does not form clear clusters in t-SNE but the classes are still linearly separable, meaning that the feature is encoded.
The last image shows two clusters but not those that separate the two classes.

A plethora of work uses t-SNE to show clusters of vectors grouped by features \citep{chi2020finding,nigam2020skillbert,wu2020tod,zhang2021mg,subakti2022performance}, though some follow-up with reporting classifier performance.
Because t-SNE visual separation is not easily quantifiable, the negative results are often underreported \citep{fanelli2012negative,mlinaric2017dealing}.
This issue can be resolved by using other methods, such as probes.

\begin{algorithm}
\caption{Forcing t-SNE clusters}
\label{alg:force_tsne}
\begin{algorithmic}
\State \Comment{Vectors of sentences in the two classes}
\State Load $D_A$, $D_B$
\State \Comment{Cluster size, e.g. $|D_A|/2$}
\State Input $c'$, $c \leftarrow c'/2$

\State \Comment{Two seeds from classes, most distant}
\State $s_A, s_B \leftarrow \arg \max_{v_A \in D_A, v_B \in D_B} ||v_A - v_B||$
\State
\State \Comment{Closest points to own seeds}
\State $C'_A \leftarrow \text{top-}c\,_{v \in D_A} \quad {-}||s_A - v||$
\State $C'_B \leftarrow \text{top-}c\,_{v \in D_B} \quad{-}||s_B - v||$
\State \Comment{Furthest points to opposing seeds}
\State $C''_A \leftarrow \text{top-}c\,_{v \in D_A} \quad||s_B - v||$
\State $C''_B \leftarrow \text{top-}c\,_{v \in D_B} \quad||s_A - v||$
\State
\State $C_A \leftarrow C'_A \cup C''_A$
\State $C_B \leftarrow C'_B \cup C''_B$
\State $\text{t-SNE}(C_A \cup C_B)$
\end{algorithmic}
\end{algorithm}

\subsection{Forcing t-SNE Clusters.}
\label{subsubsec:force_tsne}

It is possible to start with sentence vectors that result in a t-SNE graph that does not show any visual clusters and select half of them such that running t-SNE will show clusters between the two classes.
The algorithm is described in \Cref{alg:force_tsne}.
It is based on first finding two most distant ``seeds'' from the two classes and then selecting vectors of the same class which are closest to the seed or most distant to the other seed.

An example is shown in \Cref{fig:tsne_force}.
While the original does not show any clusters between the classes, the application of the algorithm selects such vectors that t-SNE shows visual clusters.
Simplu randomly subsampling the vectors would not work but this shows that using t-SNE to visually determine the presence of a feature is not robust.

\section{Conclusion}

In this work, we showed how large pre-trained language models represent sentence ambiguity in a much less extractable way than sentence complexity and stress the importance of using reasonable baselines.
We document that template-based datasets, such as BLiMP used for sentence acceptability, are not suitable for probing because of surface-level artefacts and more datasets should be developed for probing more performant models.
Finally, we discuss why using t-SNE visually for determining whether some representations contain a specific feature is not always a suitable approach.

\subsection*{Future work}

Because both t-SNE clustering and classification (inability to establish a rigid threshold for accuracy) can fail for determining whether a specific feature is represented in the model, more robust methods for this task should be devised.
These probes should also be replicated in models used for machine translation, which is the primary motivation for studying these traits.

\section{Acknowledgements}
This work has been funded from the 19-26934X (NEUREM3) grant of the Czech Science Foundation. 
The work has also been supported by the Ministry of Education, Youth and Sports of the Czech Republic, Project No. LM2018101 LINDAT/CLARIAH-CZ.

\clearpage

\bibliography{misc/bibliography}

\begin{thebibliography}{90}
\expandafter\ifx\csname natexlab\endcsname\relax\def\natexlab#1{#1}\fi

\bibitem[{Adi et~al.(2016)Adi, Kermany, Belinkov, Lavi, and
  Goldberg}]{adi2016fine}
Yossi Adi, Einat Kermany, Yonatan Belinkov, Ofer Lavi, and Yoav Goldberg. 2016.
\newblock Fine-grained analysis of sentence embeddings using auxiliary
  prediction tasks.

\bibitem[{Altmann(1985)}]{altmann1985resolution}
Gerry Altmann. 1985.
\newblock The resolution of local syntactic ambiguity by the human sentence
  processing mechanism.
\newblock In \emph{Second Conference of the European Chapter of the Association
  for Computational Linguistics}.

\bibitem[{Ambati et~al.(2016)Ambati, Reddy, and Steedman}]{ambati2016assessing}
Bharat~Ram Ambati, Siva Reddy, and Mark Steedman. 2016.
\newblock Assessing relative sentence complexity using an incremental ccg
  parser.
\newblock In \emph{HLT-NAACL}, pages 1051--1057.

\bibitem[{Baker et~al.(1994)Baker, Franz, Jordan, Mitamura, and
  Nyberg}]{baker1994coping}
Kathryn Baker, Alexander Franz, Pamela Jordan, Teruko Mitamura, and Eric
  Nyberg. 1994.
\newblock Coping with ambiguity in a large-scale machine translation system.
\newblock In \emph{COLING 1994 Volume 1: The 15th International Conference on
  Computational Linguistics}.

\bibitem[{Belinkov(2021)}]{belinkov2021probing}
Yonatan Belinkov. 2021.
\newblock Probing classifiers: Promises, shortcomings, and alternatives.

\bibitem[{Belinkov and Glass(2019)}]{belinkov2019analysis}
Yonatan Belinkov and James Glass. 2019.
\newblock Analysis methods in neural language processing: A survey.
\newblock \emph{Transactions of the Association for Computational Linguistics},
  7:49--72.

\bibitem[{Bhattacharya et~al.(2022)Bhattacharya, Kloudov{\'a}, Zouhar, and
  Bojar}]{sunit2022et}
Sunit Bhattacharya, V{\v{e}}ra Kloudov{\'a}, Vil{\'e}m Zouhar, and Ond{\v{r}}ej
  Bojar. 2022.
\newblock {EMMT}: A simultaneous eye-tracking, 4-electrode eeg and audio corpus
  for multi-modal reading and translation scenarios.
\newblock \emph{arXiv preprint arXiv:2204.02905}.

\bibitem[{Bordes et~al.(2019)Bordes, Zablocki, Soulier, Piwowarski, and
  Gallinari}]{bordes2019incorporating}
Patrick Bordes, Eloi Zablocki, Laure Soulier, Benjamin Piwowarski, and Patrick
  Gallinari. 2019.
\newblock Incorporating visual semantics into sentence representations within a
  grounded space.
\newblock In \emph{Proceedings of the 2019 Conference on Empirical Methods in
  Natural Language Processing and the 9th International Joint Conference on
  Natural Language Processing (EMNLP-IJCNLP)}, pages 696--707.

\bibitem[{Braze(2002)}]{braze2002grammaticality}
Forrest~David Braze. 2002.
\newblock \emph{Grammaticality, acceptability and sentence processing: A
  psycholinguistic study}.
\newblock University of Connecticut.

\bibitem[{Brunato et~al.(2016)Brunato, Cimino, Dell’Orletta, and
  Venturi}]{brunato2016paccss}
Dominique Brunato, Andrea Cimino, Felice Dell’Orletta, and Giulia Venturi.
  2016.
\newblock Paccss-it: A parallel corpus of complex-simple sentences for
  automatic text simplification.
\newblock In \emph{Proceedings of the 2016 Conference on Empirical Methods in
  Natural Language Processing}, pages 351--361.

\bibitem[{Brunato et~al.(2018)Brunato, De~Mattei, Dell’Orletta, Iavarone, and
  Venturi}]{brunato2018sentence}
Dominique Brunato, Lorenzo De~Mattei, Felice Dell’Orletta, Benedetta
  Iavarone, and Giulia Venturi. 2018.
\newblock Is this sentence difficult? do you agree?
\newblock In \emph{Proceedings of the 2018 Conference on Empirical Methods in
  Natural Language Processing}, pages 2690--2699.

\bibitem[{Campolungo et~al.(2022)Campolungo, Martelli, Saina, and
  Navigli}]{campolungo2022dibimt}
Niccol{\`o} Campolungo, Federico Martelli, Francesco Saina, and Roberto
  Navigli. 2022.
\newblock Dibimt: A novel benchmark for measuring word sense disambiguation
  biases in machine translation.
\newblock In \emph{Proceedings of the 60th Annual Meeting of the Association
  for Computational Linguistics (Volume 1: Long Papers)}, pages 4331--4352.

\bibitem[{Chen et~al.(2020)Chen, Fu, Xu, Xie, Tan, Chen, and
  Jing}]{chen2020probing}
Boli Chen, Yao Fu, Guangwei Xu, Pengjun Xie, Chuanqi Tan, Mosha Chen, and
  Liping Jing. 2020.
\newblock Probing bert in hyperbolic spaces.
\newblock In \emph{International Conference on Learning Representations}.

\bibitem[{Chi et~al.(2020)Chi, Hewitt, and Manning}]{chi2020finding}
Ethan~A Chi, John Hewitt, and Christopher~D Manning. 2020.
\newblock Finding universal grammatical relations in multilingual bert.
\newblock In \emph{Proceedings of the 58th Annual Meeting of the Association
  for Computational Linguistics}, pages 5564--5577.

\bibitem[{de~Vries and Nissim(2021)}]{de2020good}
Wietse de~Vries and Malvina Nissim. 2021.
\newblock As good as new. how to successfully recycle english {GPT}-2 to make
  models for other languages.
\newblock In \emph{Findings of the Association for Computational Linguistics:
  ACL-IJCNLP 2021}, pages 836--846.

\bibitem[{Devlin et~al.(2019)Devlin, Chang, Lee, and
  Toutanova}]{devlin2019bert}
Jacob Devlin, Ming-Wei Chang, Kenton Lee, and Kristina Toutanova. 2019.
\newblock Bert: Pre-training of deep bidirectional transformers for language
  understanding.
\newblock In \emph{Proceedings of the 2019 Conference of the North American
  Chapter of the Association for Computational Linguistics: Human Language
  Technologies, Volume 1 (Long and Short Papers)}, pages 4171--4186.

\bibitem[{Do et~al.(2020)Do, Zeng, and Paik}]{do2020resolving}
Quang-Minh Do, Kungan Zeng, and Incheon Paik. 2020.
\newblock Resolving lexical ambiguity in english-japanese neural machine
  translation.
\newblock In \emph{2020 3rd Artificial Intelligence and Cloud Computing
  Conference}, pages 46--51.

\bibitem[{Dong et~al.(2017)Dong, Zhang, and Yang}]{dong2017attention}
Fei Dong, Yue Zhang, and Jie Yang. 2017.
\newblock Attention-based recurrent convolutional neural network for automatic
  essay scoring.
\newblock In \emph{Proceedings of the 21st conference on computational natural
  language learning (CoNLL 2017)}, pages 153--162.

\bibitem[{{Elliott} et~al.(2016){Elliott}, {Frank}, {Sima'an}, and
  {Specia}}]{elliott-EtAl:2016:VL16}
D.~{Elliott}, S.~{Frank}, K.~{Sima'an}, and L.~{Specia}. 2016.
\newblock Multi30k: Multilingual english-german image descriptions.
\newblock In \emph{Proceedings of the 5th Workshop on Vision and Language},
  pages 70--74.

\bibitem[{Elliott et~al.(2017)Elliott, Frank, Barrault, Bougares, and
  Specia}]{elliott2017findings}
Desmond Elliott, Stella Frank, Lo{\"\i}c Barrault, Fethi Bougares, and Lucia
  Specia. 2017.
\newblock Findings of the second shared task on multimodal machine translation
  and multilingual image description.
\newblock In \emph{Proceedings of the Second Conference on Machine
  Translation}, pages 215--233.

\bibitem[{Fanelli(2012)}]{fanelli2012negative}
Daniele Fanelli. 2012.
\newblock Negative results are disappearing from most disciplines and
  countries.
\newblock \emph{Scientometrics}, 90:891--904.

\bibitem[{Foltz et~al.(1999)Foltz, Laham, and Landauer}]{foltz1999automated}
Peter~W Foltz, Darrell Laham, and Thomas~K Landauer. 1999.
\newblock Automated essay scoring: Applications to educational technology.
\newblock In \emph{Edmedia+ innovate learning}, pages 939--944. Association for
  the Advancement of Computing in Education (AACE).

\bibitem[{Hamon and Rajman(2006)}]{hamon2006x}
Olivier Hamon and Martin Rajman. 2006.
\newblock X-score: Automatic evaluation of machine translation grammaticality.
\newblock In \emph{LREC}, pages 155--160.

\bibitem[{Hanna and Bojar(2021)}]{hanna2021fine}
Michael Hanna and Ond{\v{r}}ej Bojar. 2021.
\newblock A fine-grained analysis of bertscore.
\newblock In \emph{Proceedings of the Sixth Conference on Machine Translation},
  pages 507--517.

\bibitem[{Heilman et~al.(2014)Heilman, Cahill, Madnani, Lopez, Mulholland, and
  Tetreault}]{heilman2014predicting}
Michael Heilman, Aoife Cahill, Nitin Madnani, Melissa Lopez, Matthew
  Mulholland, and Joel~R Tetreault. 2014.
\newblock Predicting grammaticality on an ordinal scale.
\newblock In \emph{ACL (2)}.

\bibitem[{Heinz and Idsardi(2011)}]{heinz2011sentence}
Jeffrey Heinz and William Idsardi. 2011.
\newblock Sentence and word complexity.
\newblock \emph{Science}, 333(6040):295--297.

\bibitem[{Hewitt et~al.(2021)Hewitt, Ethayarajh, Liang, and
  Manning}]{hewitt2021conditional}
John Hewitt, Kawin Ethayarajh, Percy Liang, and Christopher~D Manning. 2021.
\newblock Conditional probing: measuring usable information beyond a baseline.
\newblock In \emph{Proceedings of the 2021 Conference on Empirical Methods in
  Natural Language Processing}, pages 1626--1639.

\bibitem[{Hewitt and Manning(2019)}]{hewitt2019structural}
John Hewitt and Christopher~D Manning. 2019.
\newblock A structural probe for finding syntax in word representations.
\newblock In \emph{Proceedings of the 2019 Conference of the North American
  Chapter of the Association for Computational Linguistics: Human Language
  Technologies, Volume 1 (Long and Short Papers)}, pages 4129--4138.

\bibitem[{Higinbotham(1991)}]{higinbotham1991resolution}
Dan~W Higinbotham. 1991.
\newblock The resolution of lexical ambiguity in machine translation.
\newblock In \emph{Deseret Language and Linguistic Society Symposium},
  volume~17, page~7.

\bibitem[{Hoyt and Owen(2021)}]{hoyt2021probing}
Christopher~R Hoyt and Art~B Owen. 2021.
\newblock Probing neural networks with t-sne, class-specific projections and a
  guided tour.
\newblock \emph{arXiv preprint arXiv:2107.12547}.

\bibitem[{Iavarone et~al.(2021)Iavarone, Brunato, and
  Dell’Orletta}]{iavarone2021sentence}
Benedetta Iavarone, Dominique Brunato, and Felice Dell’Orletta. 2021.
\newblock Sentence complexity in context.
\newblock In \emph{Proceedings of the Workshop on Cognitive Modeling and
  Computational Linguistics}, pages 186--199.

\bibitem[{Jawahar et~al.(2019)Jawahar, Sagot, and Seddah}]{jawahar2019does}
Ganesh Jawahar, Beno{\^\i}t Sagot, and Djam{\'e} Seddah. 2019.
\newblock What does {BERT} learn about the structure of language?
\newblock In \emph{ACL 2019-57th Annual Meeting of the Association for
  Computational Linguistics}.

\bibitem[{Jin et~al.(2019)Jin, Dhingra, Cohen, and Lu}]{jin2019probing}
Qiao Jin, Bhuwan Dhingra, William Cohen, and Xinghua Lu. 2019.
\newblock Probing biomedical embeddings from language models.
\newblock In \emph{Proceedings of the 3rd Workshop on Evaluating Vector Space
  Representations for NLP}, pages 82--89.

\bibitem[{Just et~al.(1996)Just, Carpenter, Keller, Eddy, and
  Thulborn}]{just1996brain}
Marcel~Adam Just, Patricia~A Carpenter, Timothy~A Keller, William~F Eddy, and
  Keith~R Thulborn. 1996.
\newblock Brain activation modulated by sentence comprehension.
\newblock \emph{Science}, 274(5284):114--116.

\bibitem[{Kalchbrenner and Blunsom(2013)}]{kalchbrenner2013recurrent}
Nal Kalchbrenner and Phil Blunsom. 2013.
\newblock Recurrent convolutional neural networks for discourse
  compositionality.
\newblock In \emph{Proceedings of the Workshop on Continuous Vector Space
  Models and their Compositionality}, pages 119--126.

\bibitem[{Kassner et~al.(2020)Kassner, Krojer, and
  Sch{\"u}tze}]{kassner2020pretrained}
Nora Kassner, Benno Krojer, and Hinrich Sch{\"u}tze. 2020.
\newblock Are pretrained language models symbolic reasoners over knowledge?
\newblock In \emph{Proceedings of the 24th Conference on Computational Natural
  Language Learning}, pages 552--564.

\bibitem[{Kuznetsov and Gurevych(2020)}]{kuznetsov2020matter}
Ilia Kuznetsov and Iryna Gurevych. 2020.
\newblock A matter of framing: The impact of linguistic formalism on probing
  results.
\newblock In \emph{Proceedings of the 2020 Conference on Empirical Methods in
  Natural Language Processing (EMNLP)}, pages 171--182.

\bibitem[{Landauer(2003)}]{landauer2003automatic}
Thomas~K Landauer. 2003.
\newblock Automatic essay assessment.
\newblock \emph{Assessment in education: Principles, policy \& practice},
  10(3):295--308.

\bibitem[{Lavie and Denkowski(2009)}]{lavie2009meteor}
Alon Lavie and Michael~J Denkowski. 2009.
\newblock The meteor metric for automatic evaluation of machine translation.
\newblock \emph{Machine translation}, 23(2):105--115.

\bibitem[{Li et~al.(2021)Li, Zhu, Thomas, Xu, and Rudzicz}]{li2021bert}
Bai Li, Zining Zhu, Guillaume Thomas, Yang Xu, and Frank Rudzicz. 2021.
\newblock How is {BERT} surprised? {L}ayerwise detection of linguistic
  anomalies.
\newblock In \emph{Proceedings of the 59th Annual Meeting of the Association
  for Computational Linguistics and the 11th International Joint Conference on
  Natural Language Processing (Volume 1: Long Papers)}, pages 4215--4228.

\bibitem[{Lin et~al.(2014)Lin, Maire, Belongie, Hays, Perona, Ramanan,
  Doll{\'a}r, and Zitnick}]{lin2014microsoft}
Tsung-Yi Lin, Michael Maire, Serge Belongie, James Hays, Pietro Perona, Deva
  Ramanan, Piotr Doll{\'a}r, and C~Lawrence Zitnick. 2014.
\newblock Microsoft coco: Common objects in context.
\newblock In \emph{European conference on computer vision}, pages 740--755.
  Springer.

\bibitem[{Lin et~al.(2019)Lin, Tan, and Frank}]{lin2019open}
Yongjie Lin, Yi~Chern Tan, and Robert Frank. 2019.
\newblock Open sesame: Getting inside bert’s linguistic knowledge.
\newblock In \emph{Proceedings of the 2019 ACL Workshop BlackboxNLP: Analyzing
  and Interpreting Neural Networks for NLP}, pages 241--253.

\bibitem[{Liu et~al.(2019)Liu, Gardner, Belinkov, Peters, and
  Smith}]{liu2019linguistic}
Nelson~F Liu, Matt Gardner, Yonatan Belinkov, Matthew~E Peters, and Noah~A
  Smith. 2019.
\newblock Linguistic knowledge and transferability of contextual
  representations.
\newblock In \emph{Proceedings of the 2019 Conference of the North American
  Chapter of the Association for Computational Linguistics: Human Language
  Technologies, Volume 1 (Long and Short Papers)}, pages 1073--1094.

\bibitem[{Liu et~al.(2021)Liu, Zheng, Du, Ding, Qian, Yang, and
  Tang}]{liu2021gpt}
Xiao Liu, Yanan Zheng, Zhengxiao Du, Ming Ding, Yujie Qian, Zhilin Yang, and
  Jie Tang. 2021.
\newblock {GPT} understands, too.
\newblock \emph{arXiv preprint arXiv:2103.10385}.

\bibitem[{Lu et~al.(2020)Lu, Du, and Nie}]{lu2020vgcn}
Zhibin Lu, Pan Du, and Jian-Yun Nie. 2020.
\newblock Vgcn-bert: augmenting bert with graph embedding for text
  classification.
\newblock In \emph{European Conference on Information Retrieval}, pages
  369--382. Springer.

\bibitem[{Ma et~al.(2019)Ma, Wang, Ng, Nallapati, and Xiang}]{ma2019universal}
Xiaofei Ma, Zhiguo Wang, Patrick Ng, Ramesh Nallapati, and Bing Xiang. 2019.
\newblock Universal text representation from {BERT}: An empirical study.
\newblock \emph{arXiv preprint arXiv:1910.07973}.

\bibitem[{Martinc et~al.(2021)Martinc, Pollak, and
  Robnik-{\v{S}}ikonja}]{martinc2021supervised}
Matej Martinc, Senja Pollak, and Marko Robnik-{\v{S}}ikonja. 2021.
\newblock Supervised and unsupervised neural approaches to text readability.
\newblock \emph{Computational Linguistics}, 47(1):141--179.

\bibitem[{McNamara et~al.(2002)McNamara, Louwerse, and
  Graesser}]{mcnamara2002coh}
Danielle~S McNamara, Max~M Louwerse, and Arthur~C Graesser. 2002.
\newblock Coh-metrix: Automated cohesion and coherence scores to predict text
  readability and facilitate comprehension.
\newblock Technical report, Technical report, Institute for Intelligent
  Systems, University of Memphis~….

\bibitem[{Meister et~al.(2021)Meister, Pimentel, Haller, J{\"a}ger, Cotterell,
  and Levy}]{meister2021revisiting}
Clara Meister, Tiago Pimentel, Patrick Haller, Lena J{\"a}ger, Ryan Cotterell,
  and Roger Levy. 2021.
\newblock Revisiting the uniform information density hypothesis.
\newblock In \emph{Proceedings of the 2021 Conference on Empirical Methods in
  Natural Language Processing}, pages 963--980.

\bibitem[{Meyer and Lewis(2020)}]{meyer2020modelling}
Francois Meyer and Martha Lewis. 2020.
\newblock Modelling lexical ambiguity with density matrices.
\newblock In \emph{Proceedings of the 24th Conference on Computational Natural
  Language Learning}, pages 276--290.

\bibitem[{Mirault and Grainger(2020)}]{mirault2020time}
Jonathan Mirault and Jonathan Grainger. 2020.
\newblock On the time it takes to judge grammaticality.
\newblock \emph{Quarterly Journal of Experimental Psychology},
  73(9):1460--1465.

\bibitem[{Mlinari{\'c} et~al.(2017)Mlinari{\'c}, Horvat, and
  {\v{S}}upak~Smol{\v{c}}i{\'c}}]{mlinaric2017dealing}
Ana Mlinari{\'c}, Martina Horvat, and Vesna {\v{S}}upak~Smol{\v{c}}i{\'c}.
  2017.
\newblock Dealing with the positive publication bias: Why you should really
  publish your negative results.
\newblock \emph{Biochemia medica}, 27(3):447--452.

\bibitem[{Mosbach et~al.(2020)Mosbach, Khokhlova, Hedderich, and
  Klakow}]{mosbach2020interplay}
Marius Mosbach, Anna Khokhlova, Michael~A Hedderich, and Dietrich Klakow. 2020.
\newblock On the interplay between fine-tuning and sentence-level probing for
  linguistic knowledge in pre-trained transformers.
\newblock In \emph{Proceedings of the Third BlackboxNLP Workshop on Analyzing
  and Interpreting Neural Networks for NLP}, pages 68--82.

\bibitem[{Nagata(1992)}]{nagata1992anchoring}
Hiroshi Nagata. 1992.
\newblock Anchoring effects in judging grammaticality of sentences.
\newblock \emph{Perceptual and Motor Skills}, 75(1):159--164.

\bibitem[{Navigli(2009)}]{navigli2009word}
Roberto Navigli. 2009.
\newblock Word sense disambiguation: A survey.
\newblock \emph{ACM computing surveys (CSUR)}, 41(2):1--69.

\bibitem[{Nefdt(2020)}]{nefdt2020puzzle}
Ryan~M Nefdt. 2020.
\newblock A puzzle concerning compositionality in machines.
\newblock \emph{Minds and Machines}, 30(1):47--75.

\bibitem[{Nigam et~al.(2020)Nigam, Tyagi, Tyagi, and
  Saxena}]{nigam2020skillbert}
Amber Nigam, Shikha Tyagi, Kuldeep Tyagi, and Arpan Saxena. 2020.
\newblock Skillbert:“skilling” the bert to classify skills!

\bibitem[{Papadopoulou(2005)}]{papadopoulou2005reading}
Despina Papadopoulou. 2005.
\newblock Reading-time studies of second language ambiguity resolution.
\newblock \emph{Second Language Research}, 21(2):98--120.

\bibitem[{Pedregosa et~al.(2011)Pedregosa, Varoquaux, Gramfort, Michel,
  Thirion, Grisel, Blondel, Prettenhofer, Weiss, Dubourg, Vanderplas, Passos,
  Cournapeau, Brucher, Perrot, and Duchesnay}]{scikit-learn}
F.~Pedregosa, G.~Varoquaux, A.~Gramfort, V.~Michel, B.~Thirion, O.~Grisel,
  M.~Blondel, P.~Prettenhofer, R.~Weiss, V.~Dubourg, J.~Vanderplas, A.~Passos,
  D.~Cournapeau, M.~Brucher, M.~Perrot, and E.~Duchesnay. 2011.
\newblock Scikit-learn: Machine learning in {P}ython.
\newblock \emph{Journal of Machine Learning Research}, 12:2825--2830.

\bibitem[{Pericliev(1984)}]{pericliev1984handling}
Vladimir Pericliev. 1984.
\newblock Handling syntactical ambiguity in machine translation.
\newblock In \emph{10th International Conference on Computational Linguistics
  and 22nd Annual Meeting of the Association for Computational Linguistics},
  pages 521--524.

\bibitem[{Peters et~al.(2018)Peters, Neumann, Iyyer, Gardner, Clark, Lee, and
  Zettlemoyer}]{peters-etal-2018-deep}
Matthew~E. Peters, Mark Neumann, Mohit Iyyer, Matt Gardner, Christopher Clark,
  Kenton Lee, and Luke Zettlemoyer. 2018.
\newblock \href {https://doi.org/10.18653/v1/N18-1202} {Deep contextualized
  word representations}.
\newblock In \emph{Proceedings of the 2018 Conference of the North {A}merican
  Chapter of the Association for Computational Linguistics: Human Language
  Technologies, Volume 1 (Long Papers)}, pages 2227--2237, New Orleans,
  Louisiana. Association for Computational Linguistics.

\bibitem[{Pimentel et~al.(2020)Pimentel, Valvoda, Maudslay, Zmigrod, Williams,
  and Cotterell}]{pimentel2020information}
Tiago Pimentel, Josef Valvoda, Rowan~Hall Maudslay, Ran Zmigrod, Adina
  Williams, and Ryan Cotterell. 2020.
\newblock Information-theoretic probing for linguistic structure.
\newblock In \emph{Proceedings of the 58th Annual Meeting of the Association
  for Computational Linguistics}, pages 4609--4622.

\bibitem[{Radford et~al.(2019)Radford, Wu, Child, Luan, Amodei, Sutskever
  et~al.}]{radford2019language}
Alec Radford, Jeffrey Wu, Rewon Child, David Luan, Dario Amodei, Ilya
  Sutskever, et~al. 2019.
\newblock Language models are unsupervised multitask learners.

\bibitem[{Richek(1976)}]{richek1976effect}
Margaret~A Richek. 1976.
\newblock Effect of sentence complexity on the reading comprehension of
  syntactic structures.
\newblock \emph{Journal of Educational Psychology}, 68(6):800.

\bibitem[{Riezler and Maxwell~III(2006)}]{riezler2006grammatical}
Stefan Riezler and John~T Maxwell~III. 2006.
\newblock Grammatical machine translation.
\newblock In \emph{Proceedings of the Human Language Technology Conference of
  the NAACL, Main Conference}, pages 248--255.

\bibitem[{Rogers et~al.(2020)Rogers, Kovaleva, and
  Rumshisky}]{rogers2020primer}
Anna Rogers, Olga Kovaleva, and Anna Rumshisky. 2020.
\newblock A primer in bertology: What we know about how bert works.
\newblock \emph{Transactions of the Association for Computational Linguistics},
  8:842--866.

\bibitem[{{\c{S}}ahin et~al.(2020){\c{S}}ahin, Vania, Kuznetsov, and
  Gurevych}]{csahin2020linspector}
G{\"o}zde~G{\"u}l {\c{S}}ahin, Clara Vania, Ilia Kuznetsov, and Iryna Gurevych.
  2020.
\newblock Linspector: Multilingual probing tasks for word representations.
\newblock \emph{Computational Linguistics}, 46(2):335--385.

\bibitem[{Sammer et~al.(2006)Sammer, Reiter, Soderland, Kirchhoff, and
  Etzioni}]{sammer2006ambiguity}
Marcus Sammer, Kobi Reiter, Stephen Soderland, Katrin Kirchhoff, and Oren
  Etzioni. 2006.
\newblock Ambiguity reduction for machine translation: Human-computer
  collaboration.
\newblock In \emph{Proceedings of the 7th Conference of the Association for
  Machine Translation in the Americas: Technical Papers}, pages 193--202.

\bibitem[{Sarti(2020)}]{sarti-2020-interpreting}
Gabriele Sarti. 2020.
\newblock \href {https://gsarti.com/msc-thesis/introduction.html} {Interpreting
  neural language models for linguistic complexity assessment}.
\newblock Master's thesis, University of Trieste, dec.

\bibitem[{Sarti et~al.(2021)Sarti, Brunato, and
  Dell’Orletta}]{sarti2021looks}
Gabriele Sarti, Dominique Brunato, and Felice Dell’Orletta. 2021.
\newblock That looks hard: Characterizing linguistic complexity in humans and
  language models.
\newblock In \emph{Proceedings of the Workshop on Cognitive Modeling and
  Computational Linguistics}, pages 48--60.

\bibitem[{Scott(2018)}]{scott2018translation}
Bernard Scott. 2018.
\newblock \emph{Translation, brains and the computer: A neurolinguistic
  solution to ambiguity and complexity in machine translation}, volume~2.
\newblock Springer.

\bibitem[{Stahlberg and Kumar(2022)}]{stahlberg2022jam}
Felix Stahlberg and Shankar Kumar. 2022.
\newblock Jam or cream first? modeling ambiguity in neural machine translation
  with {SCONES}.
\newblock \emph{arXiv preprint arXiv:2205.00704}.

\bibitem[{{\v{S}}tajner et~al.(2017){\v{S}}tajner, Ponzetto, and
  Stuckenschmidt}]{vstajner2017automatic}
Sanja {\v{S}}tajner, Simone~Paolo Ponzetto, and Heiner Stuckenschmidt. 2017.
\newblock Automatic assessment of absolute sentence complexity.
\newblock In \emph{Proceedings of the 26th International Joint Conference on
  Artificial Intelligence, IJCAI}, volume~17, pages 4096--4102.

\bibitem[{Stanley and Gendler~Szab{\'o}(2000)}]{stanley2000quantifier}
Jason Stanley and Zoltan Gendler~Szab{\'o}. 2000.
\newblock On quantifier domain restriction.
\newblock \emph{Mind \& Language}, 15(2-3):219--261.

\bibitem[{Stymne and Ahrenberg(2010)}]{stymne2010using}
Sara Stymne and Lars Ahrenberg. 2010.
\newblock Using a grammar checker for evaluation and postprocessing of
  statistical machine translation.
\newblock In \emph{Proceedings of the Seventh International Conference on
  Language Resources and Evaluation (LREC'10)}.

\bibitem[{Subakti et~al.(2022)Subakti, Murfi, and
  Hariadi}]{subakti2022performance}
Alvin Subakti, Hendri Murfi, and Nora Hariadi. 2022.
\newblock The performance of bert as data representation of text clustering.
\newblock \emph{Journal of big Data}, 9(1):1--21.

\bibitem[{Tenney et~al.(2019)Tenney, Das, and Pavlick}]{tenney2019bert}
Ian Tenney, Dipanjan Das, and Ellie Pavlick. 2019.
\newblock Bert rediscovers the classical nlp pipeline.
\newblock In \emph{Proceedings of the 57th Annual Meeting of the Association
  for Computational Linguistics}, pages 4593--4601.

\bibitem[{Trueswell(1996)}]{trueswell1996role}
John~C Trueswell. 1996.
\newblock The role of lexical frequency in syntactic ambiguity resolution.
\newblock \emph{Journal of memory and language}, 35(4):566--585.

\bibitem[{Vaswani et~al.(2017)Vaswani, Shazeer, Parmar, Uszkoreit, Jones,
  Gomez, Kaiser, and Polosukhin}]{vaswani2017attention}
Ashish Vaswani, Noam Shazeer, Niki Parmar, Jakob Uszkoreit, Llion Jones,
  Aidan~N Gomez, {\L}ukasz Kaiser, and Illia Polosukhin. 2017.
\newblock Attention is all you need.
\newblock In \emph{Advances in neural information processing systems}, pages
  5998--6008.

\bibitem[{Vuli{\'c} et~al.(2020)Vuli{\'c}, Ponti, Litschko, Glava{\v{s}}, and
  Korhonen}]{vulic2020probing}
Ivan Vuli{\'c}, Edoardo~Maria Ponti, Robert Litschko, Goran Glava{\v{s}}, and
  Anna Korhonen. 2020.
\newblock Probing pretrained language models for lexical semantics.
\newblock In \emph{Proceedings of the 2020 Conference on Empirical Methods in
  Natural Language Processing (EMNLP)}, pages 7222--7240.

\bibitem[{Wang et~al.(2018)Wang, Singh, Michael, Hill, Levy, and
  Bowman}]{wang2018glue}
Alex Wang, Amanpreet Singh, Julian Michael, Felix Hill, Omer Levy, and Samuel~R
  Bowman. 2018.
\newblock {GLUE}: A multi-task benchmark and analysis platform for natural
  language understanding.
\newblock \emph{EMNLP 2018}, page 353.

\bibitem[{Warstadt et~al.(2020)Warstadt, Parrish, Liu, Mohananey, Peng, Wang,
  and Bowman}]{warstadt2020blimp}
Alex Warstadt, Alicia Parrish, Haokun Liu, Anhad Mohananey, Wei Peng, Sheng-Fu
  Wang, and Samuel~R Bowman. 2020.
\newblock {BLiMP}: The benchmark of linguistic minimal pairs for english.
\newblock \emph{Transactions of the Association for Computational Linguistics},
  8:377--392.

\bibitem[{Warstadt et~al.(2019)Warstadt, Singh, and
  Bowman}]{warstadt2019neural}
Alex Warstadt, Amanpreet Singh, and Samuel~R Bowman. 2019.
\newblock Neural network acceptability judgments.
\newblock \emph{Transactions of the Association for Computational Linguistics},
  7:625--641.

\bibitem[{Weller et~al.(2020)Weller, Hildebrandt, Reznik, Challis, Tass, Snell,
  and Seppi}]{weller2020you}
Orion Weller, Jordan Hildebrandt, Ilya Reznik, Christopher Challis, E~Shannon
  Tass, Quinn Snell, and Kevin Seppi. 2020.
\newblock You don’t have time to read this: An exploration of document
  reading time prediction.
\newblock In \emph{Proceedings of the 58th Annual Meeting of the Association
  for Computational Linguistics}, pages 1789--1794.

\bibitem[{Wolf et~al.(2019)Wolf, Debut, Sanh, Chaumond, Delangue, Moi, Cistac,
  Rault, Louf, Funtowicz et~al.}]{wolf2019huggingface}
Thomas Wolf, Lysandre Debut, Victor Sanh, Julien Chaumond, Clement Delangue,
  Anthony Moi, Pierric Cistac, Tim Rault, R{\'e}mi Louf, Morgan Funtowicz,
  et~al. 2019.
\newblock Huggingface's transformers: State-of-the-art natural language
  processing.
\newblock \emph{arXiv preprint arXiv:1910.03771}.

\bibitem[{Wu et~al.(2020)Wu, Hoi, Socher, and Xiong}]{wu2020tod}
Chien-Sheng Wu, Steven~CH Hoi, Richard Socher, and Caiming Xiong. 2020.
\newblock Tod-bert: Pre-trained natural language understanding for
  task-oriented dialogue.
\newblock In \emph{Proceedings of the 2020 Conference on Empirical Methods in
  Natural Language Processing (EMNLP)}, pages 917--929.

\bibitem[{Wu and Xiong(2020)}]{wu2020probing}
Chien-Sheng Wu and Caiming Xiong. 2020.
\newblock Probing task-oriented dialogue representation from language models.
\newblock In \emph{Proceedings of the 2020 Conference on Empirical Methods in
  Natural Language Processing (EMNLP)}, pages 5036--5051.

\bibitem[{Yaghoobzadeh et~al.(2019)Yaghoobzadeh, Kann, Hazen, Agirre, and
  Sch{\"u}tze}]{yaghoobzadeh2019probing}
Yadollah Yaghoobzadeh, Katharina Kann, Timothy~J Hazen, Eneko Agirre, and
  Hinrich Sch{\"u}tze. 2019.
\newblock Probing for semantic classes: Diagnosing the meaning content of word
  embeddings.
\newblock In \emph{Proceedings of the 57th Annual Meeting of the Association
  for Computational Linguistics}, pages 5740--5753.

\bibitem[{Zhang et~al.(2021)Zhang, Wu, Yang, Wu, Yi, Hsieh, Hou, and
  Cao}]{zhang2021mg}
Xiao-Chen Zhang, Cheng-Kun Wu, Zhi-Jiang Yang, Zhen-Xing Wu, Jia-Cai Yi,
  Chang-Yu Hsieh, Ting-Jun Hou, and Dong-Sheng Cao. 2021.
\newblock Mg-bert: leveraging unsupervised atomic representation learning for
  molecular property prediction.
\newblock \emph{Briefings in bioinformatics}, 22(6):bbab152.

\bibitem[{Zou and Zou(2017)}]{zou2017understanding}
Shunpeng Zou and Xiaohui Zou. 2017.
\newblock Understanding: how to resolve ambiguity.
\newblock In \emph{International Conference on Intelligence Science}, pages
  333--343. Springer.

\end{thebibliography}
\bibliographystyle{misc/acl_natbib}

\clearpage

\appendix


\end{document}